\documentclass[letterpaper]{article}

\usepackage{natbib, alifeconf}  
\usepackage{url,hyperref,cleveref}
\usepackage{booktabs}
\usepackage{amssymb}

\usepackage{lipsum}

\newcommand\blfootnote[1]{%
  \begingroup
  \renewcommand\thefootnote{}\footnote{#1}%
  \addtocounter{footnote}{-1}%
  \endgroup
}

\title{It's Much Easier for Neural Networks to Learn  Game of Life Dynamics with the Right Activation Function: Polynomial Kolmogorov-Arnold Networks}

\author{
    Tashin Ahmed$^{1, \dagger}$, \and
    Q. Tyrell Davis$^{2, \dagger}$
    \mbox{}\\
    $^1$Independent Researcher, Japan \\
    $^2$Work completed while at Cross Labs, Cross Compass Ltd., Japan. \\
    $^{\dagger}$ Equal contribution \\
    tashinahmed@aol.com
} 

\begin{document}

\maketitle

\begin{abstract}
   Previous work has found a gap between the scale of neural networks that reliably learn Conway's Game of Life, and minimal networks capable of representing the classic cellular automaton with hard-coded parameter values. Viewing neural network learning as a search process suggests a dependence on networks large enough to contain sub-networks with lucky initializations (sometimes known as `winning tickets') that actually learn the task. In this work, we reorient our perspective from discovering Life rules as a search problem back to a learning problem, and reason that with fitting inductive biases, the problem should be much more amenable to minimal networks. We find that network variants with several alternative activation functions meaningfully outperform the default choice of Rectified Linear Units, and in particular, that a $2^{nd}$ degree polynomial activation function consistently learns Life dynamics with or without the benefit of learning neural weights. Our results provide an informative demonstration of the benefits of matching learning to the task at hand and challenge the easy default choice of scale for all problems. In particular, we advocate for the use of cellular automata as simple test domains for developing strategies that can benefit machine learning for science, physics-based deep learning, and interpretable machine learning.
\end{abstract}


Data/Code available at: \url{https://github.com/TashinAhmed/LifeNNgine}
\blfootnote{\textcopyright 2026 Tashin Ahmed and Q. Tyrell Davis. Published under a Creative Commons Attribution 4.0 International (CC BY 4.0) license.}

\begin{figure*}[!h]
    \centering
    \includegraphics[width=0.8\textwidth]{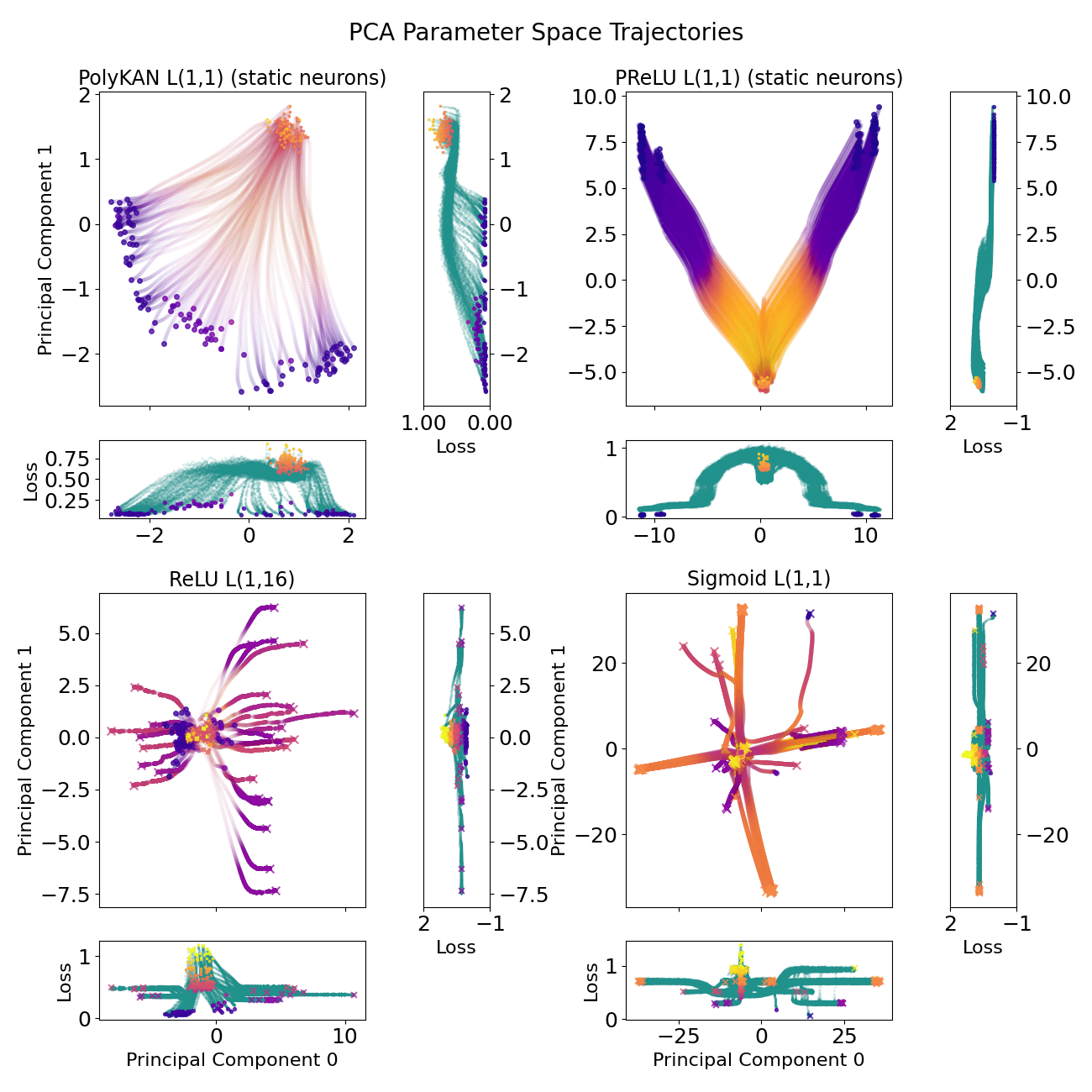}
    \caption{
        Trajectories visualized in the first two principal components of parameter space over 128 training runs for minimal PolyKAN, PReLU, ReLU, and Sigmoid networks. Side views project loss against each component. Circle/`x' end markers correspond to successful/unsuccessful training runs. Component trajectories, and start and end markers are colored light orange to dark purple according to loss. PolyKAN and PReLU are each learned by modifying activation function parameters only, ReLU and Sigmoid networks learn by adjusting synaptic weights. 
    }
    \label{fig:pca_param_space}
\end{figure*}


\section{Introduction}

The capacity of artificial Neural Networks (NNs) to model and predict complex dynamical systems is a central pursuit in the field of scientific Machine Learning (ML) and Artificial Intelligence (AI). As a tool, NNs can be used as models of complex systems and for analysis of experimental data. More broadly, world models with some physical understanding of their environment may be a necessary ingredient in developing intelligent machines. 

Cellular Automata (CA) share many traits with the structure and rules of our physical universe including locality, dynamics, and complexity. CA are widely used as models for a broad range of physical systems \citep{toffoli1987}, and the universe as cellular automaton has been an influential idea related to the concept of a computational universe of digital physics since the 1970s or earlier \citep{zuse1970}. CA are typically discrete, spatially extended dynamical systems characterized by local interactions, finite state spaces, and synchronous updates across a lattice grid. As such they can be treated as convenient `pocket universes' in exploring the expressive power, algorithmic learning capabilities, and structural inductive biases of deep NN architectures. 

Life-like CA as learning targets for convolutional NNs would appear to be a fitting match. Work in the 1980s exploring the functional approximation capabilities of NNs \citep{lapedes1987, lecun1987} eventually culminated in the universal approximation theorem (UAT), establishing NNs with a single hidden layer and nonlinear squashing function as capable of approximating arbitrary functions \citep{cybenko1989, hornik1991}. The inductive bias of 3x3 convolutional layers matches the locality defined in CA with local Moore neighborhoods, and indeed Life-like CA can be represented by and thought of as NNs with an architecture of convolutional layers followed by multilayer perceptron (MLP) layers for representing spatiality and dynamics rules, respectively \citep{gilpin2019cellular}.

Despite the approximation capabilities promised by the UAT and a close match to the inductive biases in convolutional networks, training convolutional NNs to model CA has yielded a paradoxical and highly debated observation: it is difficult for neural networks to learn the Game of Life and other Life-like CA \citep{springer2021s, gilpin2019cellular}. Kenyon and Springer found that convolutional networks with ReLU activations were unable to reliably learn the single-step dynamics of Conway's Life unless they were much wider than the minimal network required to represent the rule \citep{springer2021s}. Looking at Life-like CA more broadly, Gilpin found that networks needed to be much deeper than a minimal network, using 11 perceptron layers, to converge consistently, and found a close correspondence between the information theoretic complexity of the rule and layer activations. Both works provided hand-coded examples of minimal NNs representing Life rules: demonstrating it is not an inability of representation that makes the learning difficult.

Rectified Linear Unit (ReLU) activations are common in modern deep learning, and \citep{springer2021s} and \cite{gilpin2019cellular} both stick to this default choice. The empirical difficulty of learning Life with this standard approach highlights standard deep learning as a search process, a characterization also made by both to explain the dependence on over-parameterization. Deep learning as a search process is the foundation of the lottery ticket hypothesis \citep{frankle2018lottery} cited to explain the results in \cite{springer2021s}, entailing a dependence on scale that some, including the authors of this manuscript, find unsatisfying.

We hypothesized that taking a default choice of activation function may create an inductive bias mismatch between the weakly monotonic, piecewise linear ReLU and the interval-based logic underlying CA transition rules. We turn to the Kolmogorov-Arnold representation theorem (KART) \citep{kolmogorov1961, schmidt2021kolmogorov}, a result from real analysis that predates (and a significant influence on) the universal approximation theorem (UAT). The KART posits that any multivariate continuous function defined on a bounded domain can be exactly represented as a finite additive composition of continuous functions of a single variable \citep{kolmogorov1961}. In the context of deep learning, a resurgence of interest in the KART and its implications has motivated the ongoing development of Kolmogorov-Arnold Networks (KANs) \citep{liu2024kan}. KANs replace the synaptic weight learning paradigm of MLPs with learnable univariate activation functions, typically parameterized as a spline.

Previous work has taken inspiration from the KART and applied convolutional KANs to the problem of learning Life-like CA, notably \citep{ferenc2025}. Motivated by the KART, our approach combines recent insights in the development of KANs and polynomial neural networks \citep{ivakhnenko1971}, another idea undergoing a resurgence of interest and application in the deep learning era. Our main result is the efficacy and efficiency of a minimal model architecture with learnable polynomial activation functions,  which we call `PolyKAN'. 

PolyKAN reliably learns Game of Life using a shallower network than the 11-layer models in \citep{gilpin2019cellular} and fewer parameters than \citep{springer2021s}. Minimal PolyKAN learns Life with or without learning synaptic weights for a trainable parameter count of 34 and 29 respectively. This is comparable to a minimal, rarely converging 25 parameter ReLU network as in \citep{springer2021s}, and compares favorably to a parameter count of 140 for CKAN-C, the lowest parameter count model achieving 10/10 success reported in \citep{ferenc2025}.

Visual inspection of principal components of parameter space for training runs in \Cref{fig:pca_param_space} shows that PolyKAN trajectories appear smoother and the space of solutions to learning Life dynamics is more diverse than networks with other activation functions. Our results add to the growing case for KAN and polynomial neural networks as viable alternatives to default design choices, particularly where a mismatch exists between the dynamics being learned and the properties of ReLU NNs. This suggests a discussion of whether and when neural network learning should be viewed as a search problem where scale dominates, or as a learning problem where fitting correspondence between model and the modeled enables parameter efficiency. Conway's Life, defined by a binary state space and simple dynamics but nonetheless demonstrative of complex behavior and Turing universality \citep{berlekamp2004}, is an informative example of a `pocket universe' where the latter case is fitting.

\section{Conway's Game of Life}

Where \cite{gilpin2019cellular} trained neural networks on randomly selected CA from the space of Life-like CA and \cite{springer2021s} attempted to train NNs to predict Life states after 1 to 5 transition steps, we focus on the single-step Life problem. 

Conway’s Game of Life operates on an interval-based, strictly non-monotonic logical predicate. Let $N$ represent the total sum of living cells in the surrounding Moore neighborhood, and let $C_t$ represent the binary state of the center cell at the current time step. The definitive update rule for the cell’s state at the subsequent time step, $C_{t+1}$, is defined mathematically as follows \citep{gardner1970, berlekamp2004}:

\begin{itemize}
    \item{$C_{t + 1} = 1$, if $N = 3$. (`Birth').}
    \item{$C_{t + 1} = 1$, if $N = 3$ or $N=2$ and $C_t =1$. (`Survival')}
    \item{$C_{t + 1} = 0$, in all other cases.}
\end{itemize}

Rulestring notation defines Life-like update rules in abbreviated form, for Conway's Game the rulestring is B3/S23  (note the overlap in B/S rules at 3).  An NN must learn to approximate a highly specific square interval function: a non-monotonic mapping that remains flat at zero for inputs $N \in {0, 1}$, sharply rises to 1.0 at $N \in {2, 3}$, and remains at zero for all subsequent inputs $N \in {4, 5, 6, 7, 8}$.

\if{0}
    Despite its simplicity, Life is capable of generate complex behavior and computation \citep{berlekamp2004}, and within the family of Life-like CA (containing $262,144$ members), Life has been described as the most parsimonious CA rule exhibiting complex behavior \citep{pea2021}. 
\fi 

\section{The Lottery Ticket Hypothesis and Learning Life with ReLU networks}

\cite{springer2021s} examined the efficacy of small convolutional networks trained to predict a specified number of steps, denoted as $n$, into the future of Conway’s Game of Life. The transition rules of this two-
dimensional automaton can be implemented efficiently in a $2n + 1$ layer convolutional network requiring a minimal allocation of exactly $23n+2$ trainable parameters. 

A minimal ReLU network representing the Life transition step can be written in a few lines with parameter values given in Section A.1 by \cite{springer2021s}. Nonetheless, they reported just 3 out of 64 training instances of the minimal ReLU network learned the Life transition successfully, only achieving a  success rate in excess of 0.50 with $m=3$ overcomplete networks (69 parameters). 

A possible explanation for the failure of minimal ReLU networks can be found in the concept of the Lottery Ticket Hypothesis \citep{frankle2018lottery}. This hypothesis describes deep learning as fundamentally a lucky search problem, where dense, randomly initialized neural networks initially contain much smaller, sparse sub-networks (`winning tickets') that are able to converge via gradient descent. These sub-networks dominate model dynamics after training, whereas `losing tickets' may contribute negligibly and can be pruned away. 

According to the Lottery Ticket Hypothesis, minimal architectures fail to learn CA because learning is conceptually driven by a search over sub-networks, not present in a minimal architecture. This view is supported by sensitivity of re-training ReLU networks from previously successfully models after applying minor parameter perturbations in the addition of uniform random noise or randomly flipping parameter signs.

\subsection{The Kolmogorov-Arnold Representation Theorem}

Traditional Multi-Layer Perceptrons (MLPs) and CNNs are theoretically grounded in the Universal Approximation Theorem \citep{cybenko1989, hornik1989multilayer,hornik1991}. This theorem states that a feedforward network containing at least a single hidden layer and utilizing a fixed, non-polynomial activation function can approximate any continuous function.

KANs \citep{liu2024kan} are directly inspired by the Kolmogorov-Arnold representation theorem \citep{kolmogorov1961}. This theorem, a profound result in real analysis, posits that any multivariate continuous function defined on a bounded domain can be exactly represented as a finite composition of continuous functions of a single variable and the simple binary operation of addition. Mathematically, if $f$ is a multivariate continuous function on a bounded domain, it can be decomposed as:

\begin{equation}
f(x_1, \dots, x_n) = \sum_{q=0}^{2n} \Phi_q \left( \sum_{p=1}^n \phi_{q,p}(x_p) \right)
\end{equation}

\section{Experiments}

\begin{figure}
    \centering
    \includegraphics[width=0.275\textwidth]{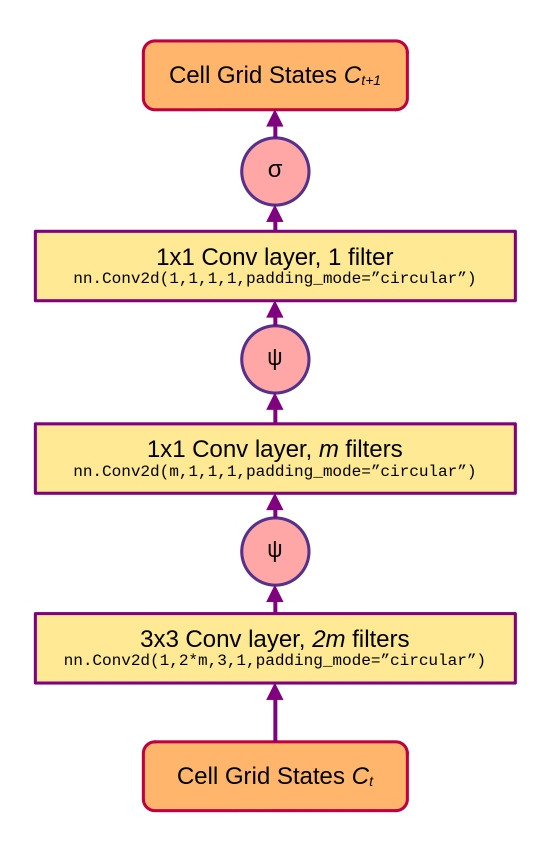}
    \caption{
        Diagram of model architecture. $\sigma$ is a sigmoid activation function, and $\psi$ represents arbitrary activation functions (including polynomials). $m$, the overcompleteness factor, modifies the width or number of filter channels in the neighborhood and update learning layers. Second line is PyTorch pseudocode used for instantiating each layer. Adapted/updated from \citep{springer2021s} Figure 2 (left). 
    }
    \label{fig:nn_diagram}
\end{figure}

\subsection{Model Framework and Training Data.}

All experiments are based on an input grid size of 32x32 cells. Grid edges are identified to make for periodic boundary conditions, and all models use `circular' padding mode to match. We matched the 8 sample batch size of \cite{springer2021s} and use a fast Life-like CA simulator written in PyTorch to continuously generate CA states from random binary initial grid states \citep{davis2021}. An epoch is not a meaningful concept if the model is not expected to ever see the same data twice, so we track model updates instead of epochs.

Model architecture is shown in \Cref{fig:nn_diagram}. Most model variants differ in the choice of activation functions readily available in PyTorch, and the polynomial activation function used in PolyKAN can be written as

\if{0}
    \begin{equation} 
        \textcolor{v0}{\phi}(\textcolor{v32}{x}) = \sum_{d=0}^
        {\textcolor{v96}{D}}
        \left(
            \textcolor{v64}{w}_d \textcolor{v32}{x}^{d}
        \right) 
    \end{equation}

    Where the polynomial function $\textcolor{v0}{\phi}$ of input $\textcolor{v32}{x}$ is defined as a sum of $\textcolor{v32}{x}$ terms exponentiated up to the function degree \textcolor{v96}{D} and weighted by coefficients $\textcolor{v64}{w}_d$. We report experiments with a 2nd degree polynomial activation function.
\fi

\begin{equation} 
    \phi({x}) = \sum_{d=0}^
    {{D}}
    \left(
        w_d x^{d}
    \right) 
\end{equation}

Where the polynomial function $\phi$ of input $x$ is defined as a sum of $x$ terms exponentiated up to the function degree $D$ and weighted by coefficients $w_d$. We report experiments with a 2nd degree polynomial activation function.

We only consider the $n=1$ single update prediction problem and therefore our models are shallow: one each 3x3 and 1x1 convolutional layers followed by a final 1x1 layer with a sigmoid activation function, $\mathcal{L}(1,m)$ in the notation of \cite{springer2021s} on which the architecture is based. Validation accuracy is calculated with a decision boundary at 0.5 with a validation set of 304 32x32 grids comprising evolved states starting from randomly initialized cell states and hand-picked gliders, oscillators, and still-life patterns. We use early stopping to truncate training when a model achieves perfect accuracy for two contiguous model updates. Hyperparameters are listed in \Cref{tabel:hyperparameters}, a summary of success rates and model variant characteristics can be found in \Cref{tab:model_success_params}.

\begin{table}
    \centering
    \begin{tabular}{cccc}
        \toprule
        Name & Value \\
        \midrule
        Optimizer  & Adam \\
        Learning Rate & 1e-3  \\
        Grid size & 32x32 \\
        Batch size & 8 \\
        Default cell density & 0.5 \\
        Convolutional filters & $2m$ \\
        1x1 update filters & $m$ \\
        update layers & $n$ \\
        \bottomrule
    \end{tabular}
    \caption{
        Hyperparameters.
    }
    \label{tabel:hyperparameters}
\end{table}

\subsection{Minimal Networks by Activation Function}

Our first experiment tested 10 different activation functions in minimal $\mathcal{L}(1,1)$ models. In addition to the custom 2nd degree polynomial used in our PolyKAN network, we tested parametric ReLU \citep{he2015}, Square ($x^2$), SiLU, RootSquare ($\sqrt{x^2}$), LeakyReLU \citep{maas2013}, CELU, Sigmoid, hyperbolic tangent, and ReLU.

\begin{figure}
    \centering
    \includegraphics[width=0.465\textwidth]{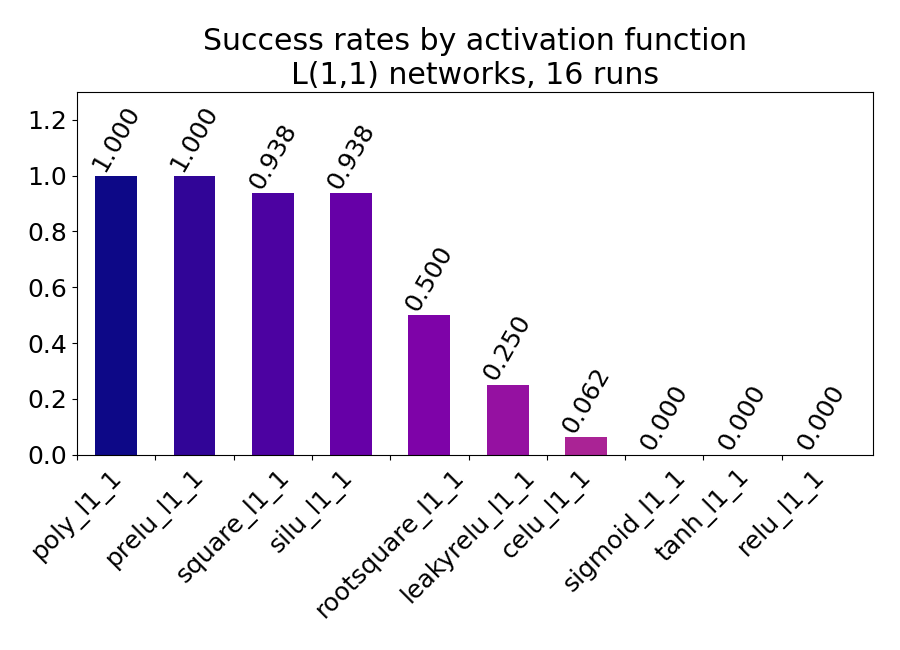}
    \caption{
        Success rates, minimal models by activation function.
    }
    \label{fig:minimal_models_by_activation}
\end{figure}

Our experiment consisted of 16 training runs and the success rates are shown in \Cref{fig:minimal_models_by_activation}. 

\subsection{Modifying Initial Density of Cell States}

A major shortcoming of training minimal ReLU conv-nets on Life is sensitivity to initial cell state density. A maximum training efficacy at about 0.38 on-density, corresponding to the maximum probability that any given cell will be on in the next time step, was previously identified \citep{springer2021s}. 

We investigated success rates for 3 of the best-performing activation functions, and ReLU, across initial on-density from 0.05 to 0.95, for 16 training runs each. We used $\mathcal{L}(1,2)$ and minimal $\mathcal{L}(1,1)$ models. 

All model variants fail entirely for initial on-state density $d_0=0.95$, where most, but not every, batch of size 8 contain no on-state cells at all after 1 step. The overall success rate of ReLU is too low to validate (or refute) the exact density optimum from \cite{springer2021s} in our relatively small number of training runs, but $\mathcal{L}(1,2)$ ReLU networks did achieve their maximum accuracy of $0.3125$ at $d_0=0.40$ and $d_0=0.60$.

\begin{figure}[!ht]
    \centering
    \includegraphics[width=0.475\textwidth]{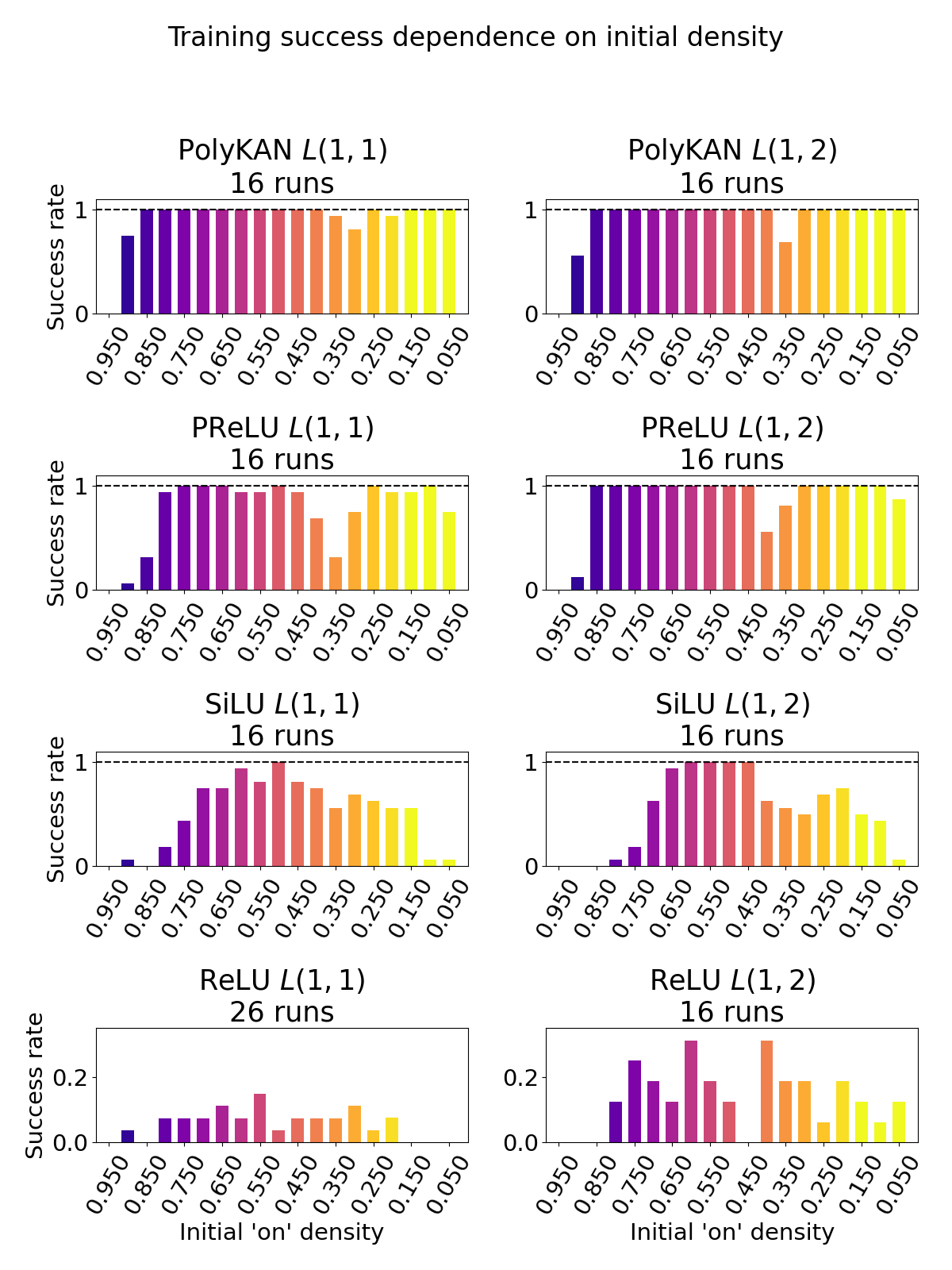}
    \caption{
        Success rates for networks with PolyKAN, PReLU, SiLU, and ReLU activation functions, subject to different initial cell state densities. Note that the y-scale for ReLU differs from the other to account for the much lower success rate.
        %
    }
    \label{fig:density}
\end{figure}

Although PolyKAN succeeds in 16/16 training runs for most initial densities, it does falter (aside from 0.0 success rate at $d_0=0.95$) at $d_0 = (0.90,0.35,0.30,0.20)$ with success rates of $(0.75,0.9375,0.8125,0.9375)$, and similarly yielded imperfect accuracies of $(0.5625,0.6875)$ at $d_0=(0.90,0.35)$ for $\mathcal{L}(1,2)$. PReLU was more acutely sensitive, with a global minimum at $0.0625$ at $d_0=0.90$ for $m=1$ and $0.1250$ at $d_0=0.90$ for $\mathcal{L}(1,1)$ and $\mathcal{L}(1,2)$. SiLU performance varied with density, with maximum success rates $1.0$ at about $d_0=0.50$. Figure \ref{fig:density} provides a visual summary of success rates by activation function initial on-state densities ranging from 0.95 to 0.05.

\subsection{Parameter Perturbation}

Even when ReLU networks are successfully trained on Life CA, small parameter perturbations significantly degrade the ability of a model to re-train on Life dynamics. The degradation applies when parameters are subjected to a number $k$ of sign flips, increasing in severity from 1 flip to 8, or the addition of uniform random noise. The degradation affects model training performance whether the perturbation is applied to final, trained parameters or to initial parameters before re-training \citep{springer2021s}. 

Following \cite{springer2021s}, we applied parameter perturbations of 0 to 8 sign flips and uniform random noise from 0.0 to 1.0 magnitude to successfully trained model parameters and to initial parameters logged from successful training runs. Due to a low number of successful $\mathcal{L}(1,1)$ ReLU networks, we used $\mathcal{L}(1,2)$.

\begin{figure}
    \centering
    \includegraphics[width=0.465\textwidth]{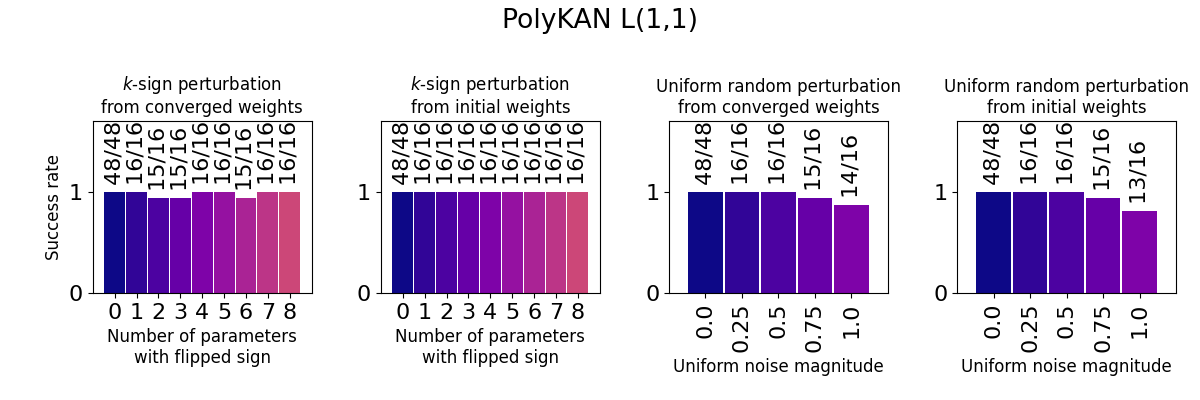}
    \includegraphics[width=0.465\textwidth]{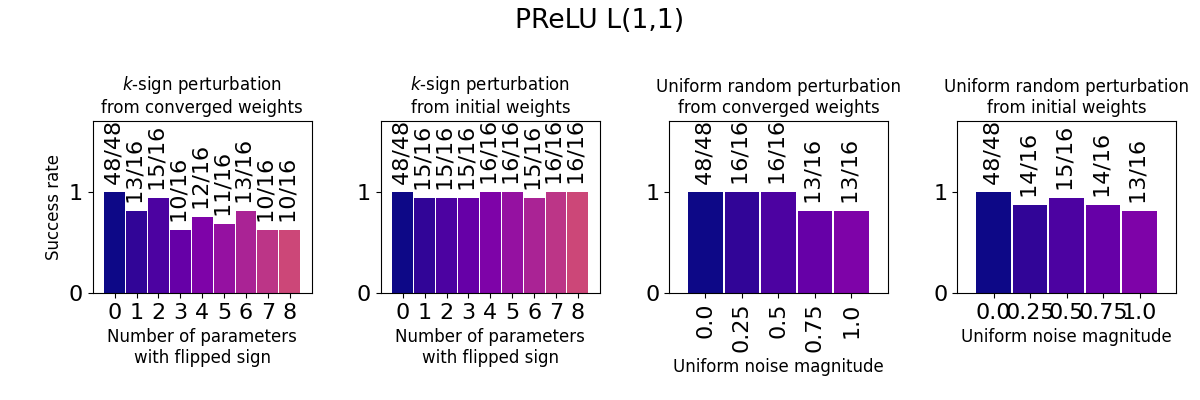}
    \includegraphics[width=0.465\textwidth]{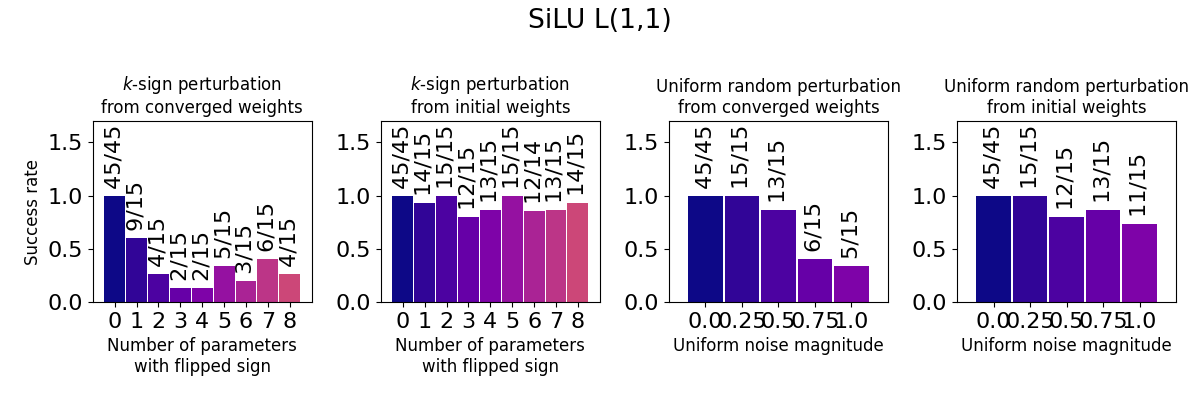}
    \includegraphics[width=0.465\textwidth]{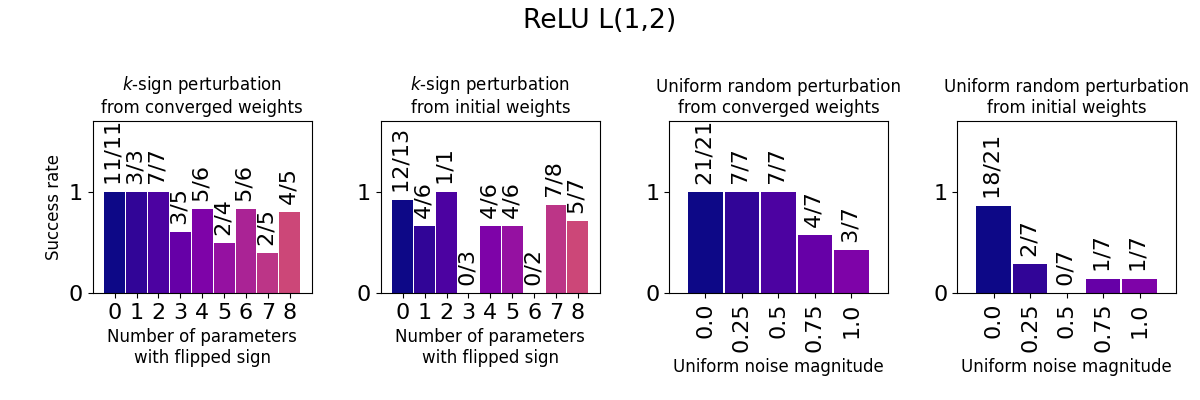}
    \caption{
        Effect of parameter perturbations on retraining success rate.
    }
    \label{fig:perturbation}
\end{figure}

Results are shown in \Cref{fig:perturbation}, with all model types demonstrating some re-training degradation with parameter perturbation. SiLU $\mathcal{L}(1,1)$ networks showed the strongest degradation of re-training performance with $k$ sign flips, though ReLU $\mathcal{L}(1,2)$ networks also failed more often with increasing $k$. PolyKAN shows some degradation re-training from final weights with $k$ sign flips (though the trend is not empirically correlated with increasing flips at these sample sizes), and notable degradation at uniform random noise perturbation at magnitudes of 0.75 and above applied to final and initial parameter values. 

\subsection{Synaptic and Activation Function Learning Ablation}

\begin{figure}
    \centering
    \includegraphics[width=0.465\textwidth]{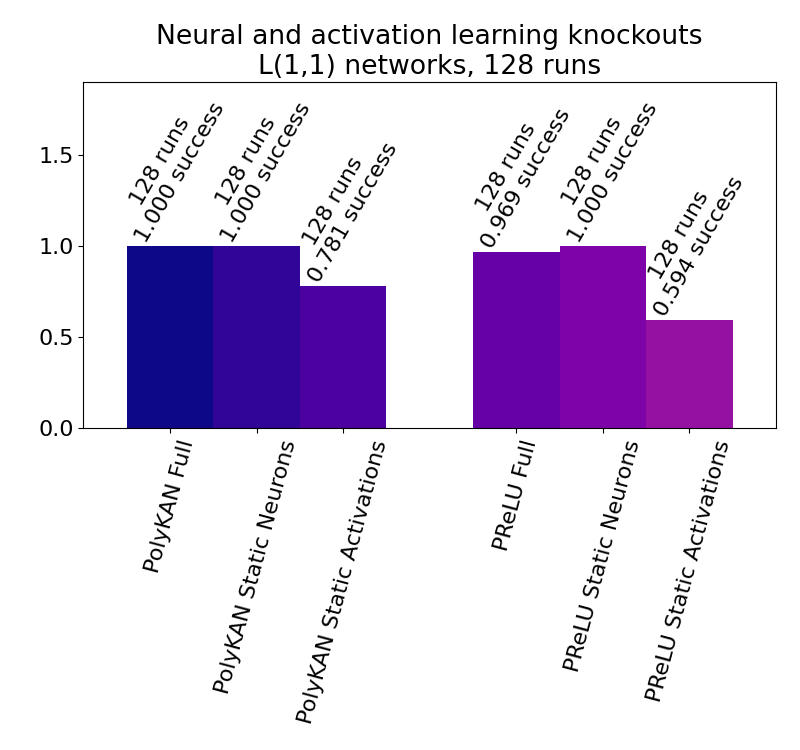}
    \caption{
        Success rates with full parameter training, static synaptic weights with activation parameters training, and synaptic weight training with static activation function parameters. Results are for $L(1,1)$ minimal models, 128 training runs for each scenario.
    }
    \label{fig:knockout_poly_prelu}
\end{figure}

Our PolyKAN implementation includes trainable synaptic weights as well as trainable coefficients in the polynomial activation functions applied to each channel. The KAN framework does not include trainable weights \citep{liu2024kan}. We performed ablation studies with either synaptic or activation function learning disabled to discover whether our PolyKAN implementation could still learn Life rules without learning any neural weights in the 1x1 update layers. We also included PReLU networks; comparisons of success rates for PolyKAN and PReLU under different ablation conditions are shown in Figure \ref{fig:knockout_poly_prelu}.

Notably the performance of PolyKAN with trainable neural weights and polynomial coefficients (34 trainable parameters) and only polynomial coefficients and 3x3 convolution weights for learning neighborhoods (29 trainable parameters) both trained successfully on Life rules in 128/128 training runs. With polynomial coefficients frozen throughout training the PolyKAN success rate dropped to 0.78 (25 trainable parameters). 

PReLU also achieved a minimal success rate, 0.59, with synaptic weight learning only (25 parameters, equivalent to a leaky ReLU with the slope for negative inputs at $a=0.25$). Over 128 runs, PReLU was successful 124 out of 128 times with full training (28 trainable parameters) and the full 128 times with update rule synaptic weights frozen (23 trainable parameters), learning only from adjusting the leaky PReLU slope.

\begin{table*}[]
\centering
\begin{tabular}{lllll}
\hline
Name       & Success Rate  & Trainable Parameters & Monotonicity    & Differentiability \\ \hline
PolyKAN    & \textbf{1.0}/\textbf{1.0}/0.78  & 34/29/25             & strictly $\rightarrow$ non    & smooth            \\
PReLU      & 0.97/\textbf{1.0}/0.59 & 28/\textbf{23}/25             & strictly $\rightarrow$ varies & discontinuous     \\
Square     & 0.94          & 25                   & non             & smooth            \\
SiLU       & 0.94          & 25                   & non             & smooth            \\
RootSquare & 0.50          & 25                   & non             & discontinuous     \\
LeakyReLU  & 0.25          & 25                   & strictly        & discontinuous     \\
CELU       & 0.06          & 25                   & strictly        & continuous        \\
Sigmoid    & 0.0           & 25                   & strictly        & smooth            \\
Tanh       & 0.0           & 25                   & strictly        & smooth            \\
ReLU       & 0.0           & 25                   & weakly          & discontinuous     \\ \hline
\end{tabular}
\caption{
        Summary of model variant characteristics and success rate learning Conway's Game of Life update rule. Numbers correspond to $\mathcal{L}(1,1)$ minimal models. PolyKAN and PReLU list performance and parameters counts training with full parameters/only activation functions/only weights modifiable. Default initializations for PolyKAN and PReLU start out strictly monotonic over the relevant numerical range, but consistently and occasionally become non-monotonic during training for PolyKAN and PReLU, respectively. PolyKAN and PReLU data are from the ablation experiment (\Cref{fig:knockout_poly_prelu}), and the others are taken from minimal networks experiment (\Cref{fig:minimal_models_by_activation}). We did not combine data from different experiments as there is some overlap in random seeds. 
    }
    \label{tab:model_success_params}
\end{table*}

\section{Discussion}

Perhaps there's no surprise in that the PolyKAN, jointly inspired by Kolmogorov-Arnold Networks and the implications of the Kolmogorov-Arnold Representation Theorem \citep{liu2024kan}, and \citep{kileel2019expressive, finkel2025activation} consistently learns the rules of Life in a minimal architecture, but that several other activation function choices do as well. In particular the performance of PReLU, little more than a glorified leaky ReLU, may be worth nothing. 

The universal approximation theorem, partly inspired by the KART, does not require that activation functions be monotonic (or not) \cite{cybenko1989, hornik1991}, though functional smoothness and smooth differentiability in the classic formulation are important \citep{hornik1990, hornik1991}. In practice ReLU NNs are capable of approximating arbitrary functions. Given the easy construction of a minimal Life NN using ReLUs, however, the smoothness and width/depth arguments \citep{shen2022} are not specifically necessary here for approximation competence. 

We note that there is no lack of capacity for minimal ReLU networks to represent Life rules perfectly. Hand-coded examples are readily available in \cite{springer2021s} and the repository for this manuscript. Instead the difficulty arises in learning, and it is there that smooth differentiability may facilitate easier success in solving Life via gradient descent. 

Parameter principal components trajectories in the loss landscape during training for several model variants are shown in \Cref{fig:pca_param_space}. Parameter component space for ReLU and Sigmoid both possess an apparent signature, potentially telling of their usual failures. Sigmoid training tends to follow one of four directions along a cross, and when viewed from the side training seems to level out as if in a local flat wash in a watershed. ReLU also appears to fall to either side of an apparent ridge, training away from a more constrained area of possible solutions near what might be `winning ticket' lucky initializations.

The difference in parameter component space during PReLU and PolyKAN training is also stark. With synaptic weight training removed, PReLU follows two general paths to a relatively small set of solution clusters bifurcated to the positive or negative side of principal component 0. PolyKAN fans out to reach a smoothly distributed frontier of good solutions, and side-view projects seem to show few barriers along the way.

\section{Conclusions}

Our results are simple: that learning single-step Game of Life dynamics is hindered by ReLU activation in minimal neural networks and significantly improved by several other activation functions, and suggestive: that choosing a fitting activation function for a given problem can improve training without falling back on an idle dependence on excessive scale. 

Taking the perspective of Life and similar CA as pocket universes with their own simple physics, the present work can be taken as a simplified analogy for physics-based deep learning and machine learning applications in science.

Among `Linear Unit' family activation functions, smoothly differentiable SiLU outperformed both discontinuously differentiable LeakyReLU and ReLU, and CELU, for which only the first derivative is continuous. Pilot experiments with other widely available LU activations seem to follow a similar pattern.

For Conway's Game of Life specifically, the comparably good success rates with minimal networks using either polynomial or PReLU activation functions, with or without synaptic weight learning, is observed in the context of two very different learnable activation functions. PReLU is piecewise linear and discontinuously differentiable, whereas the $2^{nd}$ degree polynomial functions in PolyKAN is smoothly differentiable. Neither function is constrained to be monotonic over a relevant numerical range. Further investigations examining the trajectories through loss landscapes (as in \Cref{fig:pca_param_space}) and the information theoretic diversity of their respective solution spaces may illuminate the characteristics promoting success for these variants in Life and other CA worlds.

\if{0}
    \begin{equation}
    L_1 = ReLU((C_t \circast W_{1,1} + b_{1,1}) \circled{+} (C_t \circast W_{1,2} + b_{1,2})
    \end{equation}
    \begin{equation}
    L_2 = ReLU(L_1 \circast (W_{2,1} \circled{+} W_{2,2}))
    \end{equation}
    \begin{equation}
    C_{t+1} = \sigma (L_2 \circast W_3 + b_3)
    \end{equation}
\fi

\section{Acknowledgements}

QTD and TA would like to express their sincere gratitude to the ALIFE 2026 Organizing Committee for their tremendous support, and also thanks the reviewers and meta-reviewers for their thoughtful evaluations, constructive feedback, and valuable suggestions that helped improve the quality and clarity of this work. QTD would like to thank the friendly support of colleagues at Cross Labs during the completion of this project.

\footnotesize
\bibliographystyle{apalike}
\bibliography{bibliography} 

\end{document}